\documentclass[conference]{IEEEtran}
\IEEEoverridecommandlockouts
\usepackage{cite}
\usepackage{amsmath,amssymb,amsfonts}
\usepackage{algorithm}
\usepackage{algorithmic}
\usepackage{graphicx}
\usepackage{textcomp}
\usepackage{xcolor}
\usepackage{url}
\def\BibTeX{{\rm B\kern-.05em{\sc i\kern-.025em b}\kern-.08em
    T\kern-.1667em\lower.7ex\hbox{E}\kern-.125emX}}
\begin{document}

\title{VStreamDRLS: Dynamic Graph Representation Learning with Self-Attention for Enterprise Distributed Video Streaming Solutions}

\author{\IEEEauthorblockN{Stefanos Antaris}
 \IEEEauthorblockA{KTH Royal Institute of Technology}
 \IEEEauthorblockA{Hive Streaming AB \\
 Sweden \\
 antaris@kth.se}
 \and
 \IEEEauthorblockN{Dimitrios Rafailidis}
 \IEEEauthorblockA{Maastricht University \\
 Netherlands \\
 dimitrios.rafailidis@maastrichtuniversity.nl}
}

\maketitle

\begin{abstract}
Live video streaming has become a mainstay as a standard communication solution for several enterprises worldwide. To efficiently stream high-quality live video content to a large amount of offices, companies employ distributed video streaming solutions which rely on prior knowledge of the underlying evolving enterprise network. However, such networks are highly complex and dynamic. Hence, to optimally coordinate the live video distribution, the available network capacity between viewers has to be accurately predicted. In this paper we propose a graph representation learning technique on weighted and dynamic graphs to predict the network capacity, that is the weights of connections/links between viewers/nodes. We propose VStreamDRLS, a graph neural network architecture with a self-attention mechanism to capture the evolution of the graph structure of live video streaming events. VStreamDRLS employs the graph convolutional network (GCN) model over the duration of a live video streaming event and introduces a self-attention mechanism to evolve the GCN parameters. In doing so, our model focuses on the GCN weights that are relevant to the evolution of the graph and generate the node representation, accordingly. We evaluate our proposed approach on the link prediction task on two real-world datasets, generated by enterprise live video streaming events. The duration of each event lasted an hour. The experimental results demonstrate the effectiveness of VStreamDRLS when compared with state-of-the-art strategies. Our evaluation datasets and implementation are publicly available at \url{https://github.com/stefanosantaris/vstreamdrls}.
\end{abstract}


\section{Introduction}

\begin{figure}[t!]
    \centering
    \includegraphics[scale=0.21]{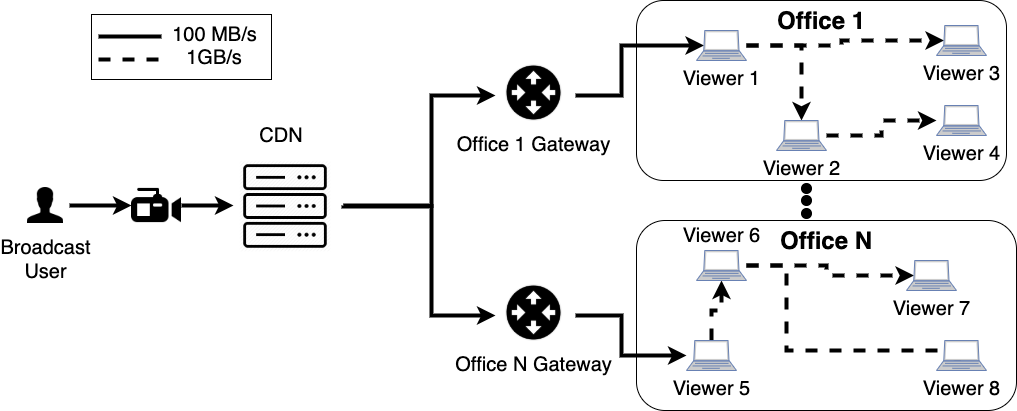}
    \caption{A distributed live video streaming process in enterprise networks. The video streaming event is produced by the broadcast user. Then, Viewers 1 and 2 are the only users who retrieve the video stream via the CDN and the respective office gateways, to distribute it to the rest of the viewers. The goal is to preserve high bandwidth connections (1GB/s) among viewers in the same office during the live video streaming event, and miminize the number of low bandwidth connections (100MB/s) among viewers in different offices.}
    \label{fig:video_distr}
\end{figure}

Live video streaming has become an essential communication solution for large organizations with several applications such as training employees, announcing product releases, and so on. For example, the fortune-500 companies\footnote{https://fortune.com/fortune500/} have several offices around the world with thousand of employees in each office. Delivering a high-quality video stream to each office is a challenging task because of the offices' network capacity limitations, and the amount of data that need to be transferred among a large number of viewers. To overcome these challenges,  enterprises apply different software solutions to transfer the video stream to each office and then distribute the video between viewers via the internal high-bandwidth network~\cite{Palacios2019,Roverso2015}. For example, as shown in Figure~\ref{fig:video_distr} each viewer is connected to a limited number of other viewers. Instead of directly connecting all viewers to the Content Delivery Network (CDN) server and download the video stream transmitted by the broadcaster, in practice only a small subset of viewers, for example, Viewer 1 and 5 in Figure~\ref{fig:video_distr} are responsible to fetch the video stream via the offices' gateways. Thereafter, each viewer exploits the connections to distribute the video stream to the remaining viewers, so as to reduce the office's network traffic and satisfy the network's capacity limitation.

To efficiently coordinate the distribution between viewers in the same office, distributed video streaming solutions require a prior knowledge of the enterprise network. Without this knowledge, the viewers may erroneously establish connections to viewers of different offices, thus significantly reducing the performance of video streaming \cite{Deng2017}. In addition, for security reasons large enterprises provide limited information about their internal network topology. A possible solution would be to exploit the network characteristics of each viewer, for example, private/public Internet Protocol (IP) addresses, and distribute the video stream between viewers with similar characteristics. However, the recent Data Protection Regulation \cite{gdpr} prohibits the collection of the IP addresses by third parties. Moreover, enterprises continuously adapt the internal network topologies, to meet the evolving requirements of their employees. To overcome the problem of the unknown and evolving internal network topology, baseline strategies for video streaming distribution such as~\cite{Edan2017,Brice2018} allow each viewer to establish random connections and periodically adapt their connections based on the observed network capacity until it converges to the underlying enterprise network. Despite their convergence guarantees, the time complexity to infer the underlying network increases exponentially along with the number of viewers in the enterprise. In practice, a large number of viewers from different offices participates in real-world enterprise live video streaming events, for example, thousands of viewers at hundred of offices. Meanwhile, the average duration of the event lasts several minutes, for instance, from $30$ to $60$ minutes. As a consequence, randomly selecting the connections that each viewer maintains has a negative impact on the performance of a video streaming event \cite{Deng2017}. Therefore, it is important to predict the network capacity between different viewers in real-time during a live video streaming event and coordinate the connections to distribute the video stream among viewers, accordingly. 

During a live video streaming event, the number of viewers and their  connections significantly vary over time, as viewers emerge and leave at unexpected pace. At the same time each viewer has to maintain a limited number of connections. A live video streaming event can be modeled as a dynamic and weighted graph, where viewers and their connections are the nodes and edges of the graph, respectively. The weight of the connection corresponds to the network capacity between two nodes. The generated graphs are highly dimensional, sparse and significantly evolve over time. To reduce the graph dimensionality, representation learning approaches have recently introduced to learn low-dimensional node features as latent representations~\cite{Grover2016,hamilton17,Perozzi2014}. However, such approaches are designed to work on static graphs, ignoring the evolution of the graph during a live video streaming event. Recent attempts on dynamic graphs extend the static approaches by enforcing smoothness techniques \cite{Goyal2018,hamilton2017,Zhou2018} or designing recurrent neural networks \cite{Goyal2020,Hajiramezanali2019,Pareja2020} between consecutive graph convolutional networks (GCNs) \cite{kipf2017}. Most recently self-attention mechanisms have been explored to learn node representations by identifying the significance of each edge between sequential graph snapshots~\cite{Sankar2020}. Although state-of-the-art dynamic approaches are effective on high-sparsity settings, they achieve low performance on graphs generated by live video streaming events. This occurs because consecutive graph snapshots significantly differ during live video streaming events as we will show in our experiments. 

To overcome the shortcomings of existing strategies in the case of video streaming distribution, in this paper we propose a neural network model Dynamic Graph Representation Learning for enterprise Video Streaming with Self-Attention, namely VstreamDRLS. Our model adapts the GCN architecture to address the problem of learning node representations on dynamic graphs. Provided that the graphs generated by live video streaming events evolve over time, we introduce a self-attention mechanism on the GCN to focus on the parameter weights that are relevant to the evolution of the graph. Our main contributions are summarized as follows: 

\begin{itemize}
    \item We demonstrate the limitations of existing graph representation learning approaches on the evolving graphs generated by video streaming distribution events. To the best of our knowledge we are the first who considered the problem of live video streaming on enterprise networks as a problem of dynamic graph representation learning.
    \item We propose a neural network model that integrates a self-attention mechanism into graph convolutional networks to learn latent node representations when graph snapshots vary significantly. In particular, when the graph  significantly changes the self-attention mechanism tends to forget the historical information and forces the GCN to produce different latent node representations, accordingly. This is achieved by performing self-attention on the  weights  of  the  GCN  parameters and capturing  the  temporal evolution accordingly. In doing so, our model captures the graph structure evolution and learns node representations to efficiently predict the network capacity between different viewers/nodes during a live video streaming event.
\end{itemize}
Our experiments on real-world datasets from video streaming distribution events demonstrate the superiority of our model over other state-of-the-art methods. 

The remainder of the paper is organized as follows: Section~\ref{sec:rel} reviews the related work and in Section~\ref{sec:prop} we detail the proposed VstreamDRLS model. Our experimental evaluation is presented in Section~\ref{sec:exp} and we conclude the study in Section~\ref{sec:conc}.

\section{Related Work}\label{sec:rel}

State-of-the-art graph representation learning techniques calculate low-dimensional latent node representations from two types of graphs: (i) static graphs where nodes and edges are fixed, and (ii) dynamic graphs where both nodes and edges evolve on different graph snapshots. Static approaches exploit a wide range of techniques to learn node representations, such as matrix factorization \cite{Cao2015,Ou2016}, Random Walks \cite{Grover2016,hamilton2017,Perozzi2014}, Deep AutoEncoders \cite{Wang16}, Graph Convolutions \cite{kipf2017}, Graph Attentions \cite{velickovic2018, Kefato2020}, and Adversarial Learning \cite{Pan2019}. Despite their success on static graphs, these methods fail to capture the evolution of dynamic graphs.

Recent approaches on dynamic graphs aim to learn the temporal dynamics over consecutive graph snapshots. For example, Dynamic Joint Variational Graph AutoEncoder (DynVGAE) shares weights between consecutive GCNs and models the graph evolution by formulating a joint loss function~\cite{mahdavi2019}. Recently, a set of approaches tries to summarize the graph evolution based on recurrent architectures such as Gated Recurrent Units (GRUs) between GCNs. For instance,  Graph Convolutional Recurrent Network (GCRN) exploits GCNs to compute the node representations and then provides the generated representations to Long-Short Term Memory (LSTM) networks to learn the graph dynamics~\cite{Seo2016}. Dyngraph2vec stacks several LSTMs in the AutoEncoder architecture to learn the long-term dependencies of the dynamic graph~\cite{Goyal2020}. Evolving Graph Convolutional Network (EvolveGCN) employs GRUs to store the importance of the node features in the hidden states and learn the weights of each GCN layer~\cite{Pareja2020}. Dynamic Self-Attention Network (DySAT) captures the graph evolution by applying a self-attention mechanism to focus on the important node features and edges that are preserved over consecutive graph snapshots~\cite{Sankar2020}. Instead our self-attention mechanism differs from DySAT, as in our architecture the proposed self-attention mechanism is designed to capture the temporal evolution on the weights of the GCN parameters and not the node representations as DySAT does. Despite the ability of dynamic approaches to learn node representations on evolving graphs, these approaches underperform in the link prediction task during live video streaming events (Section~\ref{sec:exp}).

\section{Proposed Model}\label{sec:prop}

A dynamic network $\mathcal{G}$ is defined as a sequence of graph snapshots $\mathcal{G} = \{\mathcal{G}_1, \mathcal{G}_2, \dots ,\mathcal{G}_K\}$ evolving over $K$ time steps. For each time step $k=1,\ldots,K$, the graph snapshot $\mathcal{G}_k$ is denoted by $\mathcal{G}_k = (\mathcal{V}_k, \mathcal{E}_k, \mathbf{X}_k)$, where $\mathcal{V}_k$ is the set of $n_k = |\mathcal{V}_k|$ nodes and $\mathcal{E}_k$ corresponds to the set of edges  $\mathcal{E}_k$. $\mathbf{X}_k \in \mathbb{R}^{n_k \times m}$ is the node feature matrix, where each node has $m$ features. The $k$-th graph snapshot corresponds to a weighted adjacency matrix $\mathbf{A}_k \in \mathbb{R}^{n_k \times n_k}$, with $A_k(u,v) > 0$, for nodes $u$ and $v \in \mathcal{V}_k$, if $e(u,v) \in \mathcal{E}_k$. The problem of dynamic graph representation learning is to compute the latent node representation matrix $\mathbf{Z}_k \in \mathbb{R}^{n_k \times d}$ at each time step $k=1,\ldots,K$, with $d \ll m$. The computed node representations in $\mathbf{Z}_k$ should capture the evolution of the graph up to the time step $k$, on condition that the pairwise node representations similarities approximate the adjacency matrix $\mathbf{A}_k$ \cite{hamilton17}.

\begin{figure*}
    \centering
    \includegraphics[width=13cm,height=5cm]{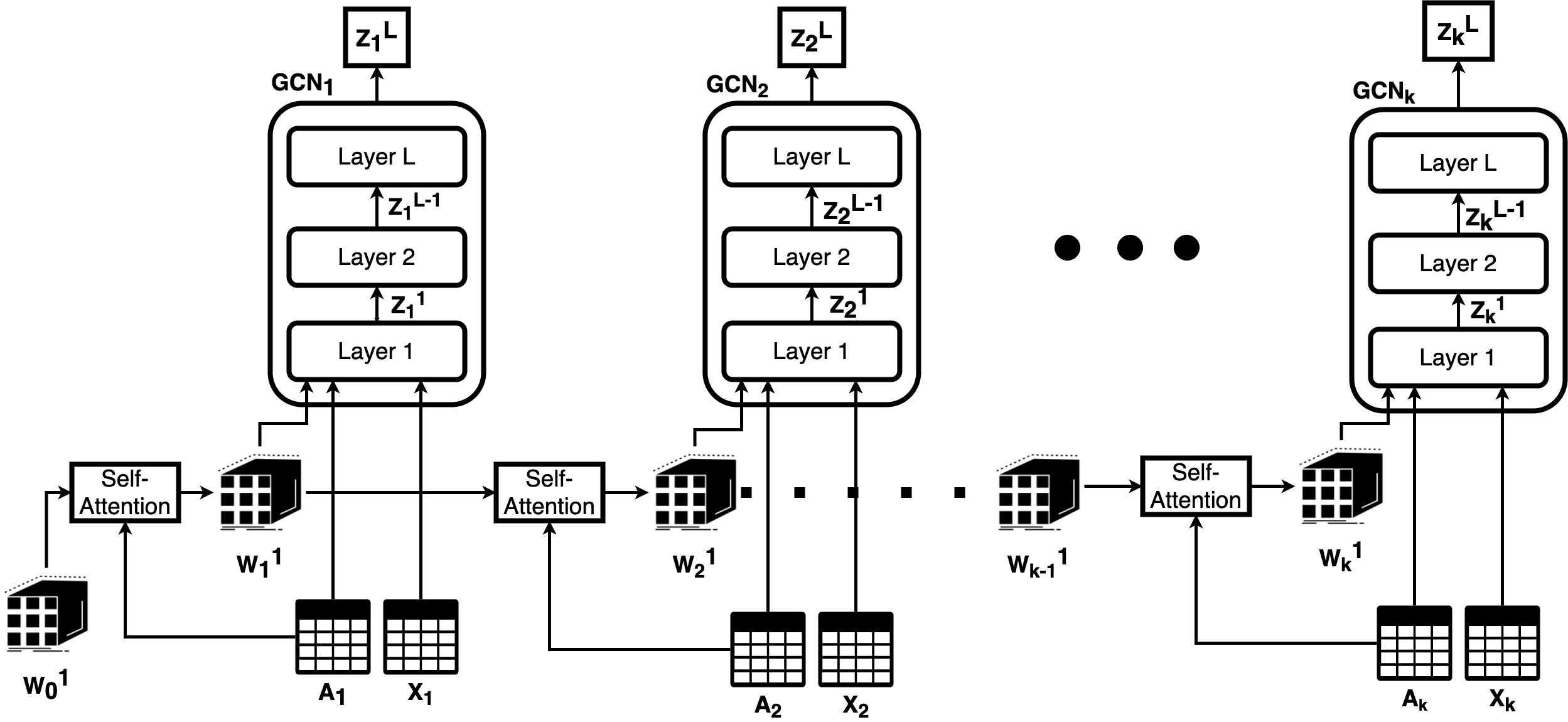}
    \caption{Overview of the architecture of the proposed VStreamDRLS model for $k$ consecutive graph snapshots. Each GCN model consists of $L$ convolutional layers. To capture the graph structure evolution, the parameter weight matrix $\mathbf{W}_k^{1}$ at the first convolutional layer of each GCN$_k$ model is calculated based on the self-attention mechanism on the parameter weight matrix $\mathbf{W}_{k-1}^1$ of the previous GCN$_{k-1}$ model and the adjacency matrix $\mathbf{A}_k$. The output of the proposed VStreamDRLS model is the final  latent node representation matrix $\mathbf{Z}_k=\mathbf{Z}^L_k$ at the top $L$-th layer of the last GCN$_k$ model.}
    \label{fig:arch}
\end{figure*}

An overview of the architecture of the proposed model is illustrated in Figure~\ref{fig:arch}. For a streaming event up to the $k$-th time step, we consider $k$ sequentially coupled GCN models. Each GCN model has $L$ convolutional layers. The inputs of the first convolutional layer of each GCN$_k$ model are (i) the adjacency matrix $\mathbf{A}_k$, (ii) the node feature matrix $\mathbf{X}_k$ and (iii) the parameter weight matrix $\mathbf{W}^1_{k-1} \in \mathbb{R}^{m \times d_1}$ at the first convolutional layer of the previous GCN$_{k-1}$ model, with $d_1$ being the dimensionality of the latent node representations produced by the first convolutional layer. To capture the graph structure evolution, the self-attention mechanism is applied on the previous parameter weight matrix $\mathbf{W}^1_{k-1}$ for calculating the parameter weight matrix $\mathbf{W}^1_{k} \in \mathbb{R}^{m \times d_1}$ at the first layer of each GCN$_{k}$ model, by taking into account how much the nodes' neighborhoods have been preserved when transitioning from time step $k-1$ to $k$. The output of the proposed model is the final latent node representation matrix $\mathbf{Z}_k=\mathbf{Z}^L_k  \in \mathbb{R}^{n_k \times d}$, generated by the top ($L$-th) layer of the last GCN$_{k}$ model, with $d=d_L$ being the dimensionality of the final latent node representations. 

In the rest of this Section we first introduce the baseline GCN model~\cite{kipf2017} on static graphs in Section~\ref{sec:GCN}, and then present the proposed model in Section~\ref{sec:VstreamDRLS}. Finally, we outline the learning strategy of our model in Section~\ref{sec:alg}. 

\subsection{Graph Convolutional Network}\label{sec:GCN}

As the baseline GCN model~\cite{kipf2017} is designed to process static graphs, for clarity we omit the time index $k$ for all the graph variables in this Section. A GCN consists of $l=1,\ldots,L$ convolutional layers, stacked sequentially as shown in Figure~\ref{fig:arch}. This means that the input of the $l$-th layer is the output of the previous $l-1$ layer. The reason for considering multiple convolutional layers is to sequentially aggregate the high-dimensional node representations to their neighborhood and embed nodes with similar representations into a significantly low-dimensional latent space~\cite{kipf2017}. Therefore, the GCN model takes as input a normalized adjacency matrix $\mathbf{\hat{A}} \in \mathbb{R}^{n \times n}$, where $n$ is the number of nodes. To compute the low-dimensional latent node representations $\mathbf{Z}^l \in \mathbb{R}^{n \times d_{l}}$ at each convolutional layer $l$, we aggregate the latent representations $\mathbf{Z}^{l-1} \in \mathbb{R}^{n \times d_{l-1}}$ of the node neighborhood at the previous $l-1$ convolutional layer, with $d_l < d_{l-1}$, as follows:

\begin{equation}
    \mathbf{Z}^l = f(\mathbf{\hat{A}} \mathbf{Z}^{l-1} \mathbf{W}^{l}) 
    \label{eq:emb}
\end{equation}
where $\mathbf{\hat{A}}$ is a symmetrically normalized adjacency matrix which is defined as:

\begin{equation}
\begin{array}{l}
    \mathbf{\hat{A}} = \mathbf{D}^{-1/2} \mathbf{\tilde{A}} \mathbf{D}^{-1/2} \\
    \mathbf{\tilde{A}} = \mathbf{A} + \mathbf{I} \\
    \mathbf{D} = \textnormal{\textbf{diag}} \big( \sum_j{A(u,v)} \big)
\end{array}
\label{eq:normalize}
\end{equation}
and $f$ is a non-linear activation function, such as $ReLU(x)=\max(0,x)$. The activation function at the last ($L$-th) layer is the linear function $f(x) = x$. In Equation~(\ref{eq:emb}), the parameter weight matrix $\mathbf{W}^l \in \mathbb{R}^{d_{l-1} \times d_l}$ computes the latent node representations $\mathbf{Z}^l$ at the $l$-th layer. At the last $L$-th convolutional layer, the parameter weight matrix $\mathbf{W}^L \in \mathbb{R}^{d_{L-1} \times d_L}$ is used to compute the final latent node representation matrix $\mathbf{Z}^L \in \mathbb{R}^{n \times d_{L}}$. For simplicity, we set $\mathbf{Z}=\mathbf{Z}^L$ and $d=d_L $. Note that the input of the first convolutional layer is the node feature matrix $\mathbf{X} \in \mathbb{R}^{n \times m}$. If the graph does not have node features, as it happens in ours graphs during the live video streaming events, the node feature matrix $\mathbf{X}$ is replaced by the identity matrix $\mathbf{I} \in \mathbb{R}^{n \times n}$, with $m=n$.  

\subsection{VstreamDRLS}\label{sec:VstreamDRLS}

The input of the self-attention mechanism in each GCN$_{k}$ model is the parameter weight matrix $\mathbf{W}^1_{k-1}$ at the first layer of the previous GCN$_{k-1}$ model and the adjacency matrix $\mathbf{A}_k$. Provided that in our setting holds $m=n_k$, thus $\mathbf{W}^1_{k-1} \in \mathbb{R}^{n_k \times d_1}$, $\forall$ node $v\in \mathcal{V}_t$ we calculate the parameter weight vector $\mathbf{W}^1_k(v)\in \mathbb{R}^{d_1}$ of the GCN$_{k}$ model as follows:

\begin{equation}
    \mathbf{W}^1_k(v) = ELU \big(\sum_{u \in \mathcal{N}_v} \alpha_{u,v} \mathbf{H}_k \mathbf{W}^1_{k-1}(u)\big)
    \label{eq:attentweight}
\end{equation}
where $\mathcal{N}_v = \{u \in \mathcal{V}_t : (v,u) \in \mathcal{E}_t\}$ is the neighborhood set of node $v$ at the graph snapshot $\mathcal{G}_t$ based on the adjacency matrix $\mathbf{A}_k$, and $ELU$ is the Exponential Linear Unit activation function. Variable $\mathbf{H}_k \in \mathbb{R}^{d_1 \times d_1}$ is the transformation matrix that needs to be learned for the weight matrices $\mathbf{W}_k^1$ and $\mathbf{W}_{k-1}^1$ at the first convolutional layers of the GCN$_{k}$ and GCN$_{k-1}$ models. To measure the importance of the neighbor $u$ to the node $v$ in Equation~(\ref{eq:attentweight}), we perform self-attention on the nodes $u$ and $v$, by applying the attention weight $\alpha_{u,v}$ to the transformed weight vector $\mathbf{H}_k \mathbf{W}^1_{k-1}(u) \in \mathbb{R}^{d_1}$, as shown in Equation~(\ref{eq:attentweight}). The attention weight $\alpha_{u,v}$ corresponds to the normalized value of the attention coefficient $c_{u,v}$,  calculated based on the softmax function\cite{velickovic2018, Sankar2020} as follows:

\begin{equation}
\begin{array}{c}
    \alpha_{u,v} = \frac{\exp (c_{u,v})}{\sum_{w\in\mathcal{N}_v} \exp (c_{w,v})} \\ \\
     c_{u,v} = \sigma \big( A_k(u,v) \cdot \mathbf{a}^\top_k [\mathbf{H}_k \mathbf{W}^1_{k-1}(u) \parallel \mathbf{H}_k \mathbf{W}^1_{k-1}(v)] \big)
\end{array}
\label{eq:softmax}
\end{equation}
where $\sigma$ is the sigmoid function, $A_k(u,v)$ is the edge weight of nodes $u$ and $v\in\mathcal{V}_k$, and $\mathbf{a} \in \mathbb{R}^{2d_1}$ is the $2d_1$-dimensional parameter vector in the self-attention mechanism. Symbol $\parallel$ denotes the concatenation operation.

Higher attention weights $\alpha_{u,v}$ reflect on more important neighbors when the graph evolves over time. This means that high attention weights correspond to neighbors that are preserved over consecutive snapshots, whereas low attention weights are computed for neighbors that are not maintained over the graph evolution. In practice, during a live video streaming event this means that high attention weights correspond to neighbors with a high network capacity over consecutive snapshots, whereas low attention weights are assigned to neighbors with a low network capacity.

In Equation (\ref{eq:softmax}), the attention coefficient $c_{u,v}$  expresses the difference between two consecutive graph snapshots. The coefficient value between node $v$ and its neighbor $u$ is decreased when the graph snapshots significantly differ. In doing so, our model forgets the previous weight vector $\mathbf{W}^1_{k-1}(v)$ when the neighbors of node $v$ significantly change. Otherwise, the attention coefficients have high values which reflect on the importance of the connection between node $v$ and its neighbor $u$. In doing so, we are able to attend the evolution of the graph and directly reflect this evolution on the parameter weight matrix  $\mathbf{W}^1_{k}$ of each GCN$_{k}$ model. 

\subsection{Learning Strategy} \label{sec:alg}

Provided $k=1,\ldots,K$ graph snapshots during a live video streaming event, in our model we have to train $k$ consecutive GCN models, simultaneously. Instead of considering all the $k$ different graph snapshots/time steps, in practice in our implementation we consider a time window $w<k$. Therefore, our model consists of $w$ consecutive GCN models $\{GCN_{k-w},\ldots, GCN_k\}$. As we will show later in Section~\ref{sec:param} considering all the $k$ different time steps does not necessarily pay off in terms of the link prediction accuracy. In doing so, we also avoid to compute $k$ different GCNs, thus highly reducing the complexity of our model. 

As aforementioned, within a time window $w$ the GCN models $\{GCN_{k-w},\ldots, GCN_k\}$ are connected in a sequential manner via the weights $\mathbf{W}^1_{k-1}$ and $\mathbf{W}^1_{k}$ at the first convolutional layers of two consequtive GCNs. Each GCN$_k$ model takes as input the adjacency matrix $\mathbf{A}_k$, the node features $\mathbf{X}_k$ and the parameter weight matrix $\mathbf{W}^1_{k-1}$ at the first convolutional layer of the previous $GCN_{k-1}$  model. Note that the parameter weight matrix $\mathbf{W}^1_{1}$ of the first GCN$_1$ model is randomly initialized. When training our model, each GCN$_k$ model  updates the respective weight matrix $\mathbf{W}^1_{k}$ according to the attention mechanism in Equation (\ref{eq:attentweight}) and then computes the final node latent representation matrix $\mathbf{Z}_k$ based on Equation~(\ref{eq:emb}). To train our model and compute the final node latent representation matrix $\mathbf{Z}_k$, we formulate the following Root Mean Square Error loss function with respect to $\mathbf{Z}_k$:

\begin{equation}
    \min_{\mathbf{Z}_k}{\mathcal{L}} = \sqrt{\frac{1}{n_k}\sum_{v \in \mathcal{V}_k}\sum_{u \in \mathcal{N}_v} {\bigg(f \big(\mathbf{Z}^\top_k(u) \mathbf{Z}_k(v)\big)  - A_k(u,v)\bigg)^{2}}}
    \label{eq:loss}
\end{equation}
where $f$ is the $ReLU$ activation function and the term \big($\mathbf{Z}^\top_k(u) \mathbf{Z}_k(v) - A_k(u,v)$\big) expresses the prediction error of how well the neighborhood $\mathcal{N}_v$ of node $v$ is preserved in the $d$-dimensional node representations $\mathbf{Z}_k(u)$ and $\mathbf{Z}_k(v)$, when compared with the connections/links in the adjacency matrix $A_k(u,v)$.

In Algorithm 1, we present the steps to learn the final latent node representations at time step $k$, given a window size $w$. The inputs of VStreamDRLS model are the $w$ adjacency matrices $\{\mathbf{A}_{k-w}, \ldots, \mathbf{A}_{k}\}$, and the feature matrices $\{\mathbf{X}_{k-w},\ldots,\mathbf{X}_{k}\}$. The output of our model is the final node latent representation matrix $\mathbf{Z}_k$. To train our model, the goal is to minimize the loss function $\mathcal{L}$ in Equation~(\ref{eq:loss}), with respect to the final node latent representation matrix $\mathbf{Z}_k$. At the beginning of the training, the $w$ different transformation matrices $\mathbf{H}_k$ are randomly initialized~\cite{Glorot2010}. In lines 2-4, we calculate the parameter weight matrix $\mathbf{W}^1_k$, also requiring to recursively compute the $w$ previous weight matrices $\mathbf{W}^1_{k-1}$, $\mathbf{W}^1_{k-2}$,$\dots$,$\mathbf{W}^1_{k-w}$ based on Equation (\ref{eq:attentweight}). In line 5, we convolute the adjacency matrix $\mathbf{A}_k$ with the weights $\mathbf{W}^1_k$ based on Equation (\ref{eq:emb}) to generate the final node latent representation matrix $\mathbf{Z}_k$. In lines 6-7, we calculate the loss function $\mathcal{L}$, and optimize the parameters ${\mathbf{H}_{k-w}, \ldots, \mathbf{H}_{k}}$ based on the backpropagation algorithm with the Adam optimizer \cite{kingma2014}. We repeat the process in lines 1-8 until the algorithm convergences, and compute the final node latent representation matrix $\mathbf{Z}_k$.

\begin{algorithm}[H]
\label{alg:vstreamdrls}
 \caption{VStreamDRLS model}
 \begin{algorithmic}[1]
 \renewcommand{\algorithmicrequire}{\textbf{Input:}}
 \renewcommand{\algorithmicensure}{\textbf{Output:}}
 \REQUIRE $\{\mathbf{A}_{k-w}, \ldots, \mathbf{A}_{k}\}$, $\{\mathbf{X}_{k-w},\ldots,\mathbf{X}_{k-w}\}$
 \ENSURE  $\mathbf{Z}_k$
 \\ \textit{Initialisation} : $\{\mathbf{H}_{k-w}, \ldots, \mathbf{H}_{k}\}$
 \REPEAT
 \FOR {($i = k-w$ to $k$)}
    \STATE $\mathbf{W}^1_i = $ \textit{CalculateWeights}($\mathbf{A}_i, \mathbf{W}^1_{i-1}, \mathbf{H}_{i})$
 \ENDFOR 
 \STATE $\mathbf{Z}_k =$ GCN$(\mathbf{A}_k, \mathbf{X}_k, \mathbf{W}^1_{k})$ 
 \STATE $\mathcal{L}$ = loss($\mathbf{Z}_k$, $\mathbf{A}_k$)
 \STATE $\{\mathbf{H}_{k-w}, \ldots, \mathbf{H}_{k}\} \leftarrow $ \textit{UpdateParameters($\mathcal{L}$)}
 \UNTIL{convergence}
 
 \RETURN $\mathbf{Z}_k$
 \end{algorithmic}
 \end{algorithm}

\section{Experiments}\label{sec:exp}

\subsection{Datasets} \label{sec:setup}

Our evaluation datasets were generated by two real live video streaming events, operated on two different enterprise networks. We refer to the generated datasets as \textbf{LiveStream-400} and \textbf{LiveStream-20K}. Both datasets are anonymized and we made the generated graphs publicly available.

\textbf{LiveStream-400} is a weighted undirected graph with $386$ viewers/nodes and $61,125$ connections/edges in total. The viewers were equally distributed to $7$ different offices around the world and each viewer attended the video streaming event for maximum an hour. The dataset consists of $12$ graph snapshots, collected every $5$ minutes during the video streaming event. To generate this dataset, we  modified the distribution software, so as to remove the connectivity limitation that each viewer/node has, which means that there was no limitation of how many connections the viewers could establish. In doing so, the number of viewers participated into the live video streaming event was on purpose limited to $400$, so that our video distribution software has no negative impact on the customer's experience in the real video streaming event. 

\textbf{LiveStream-20K} was generated based on an unmodified video distribution software provided by our company and each viewer had a connectivity limitation of maximum $7$ connections. The dataset consists of $20,357$ viewers/nodes and $812,810$ connections/edges in total over $12$ graph snapshots. The viewers were equally distributed to $25$ different offices and as in the previous dataset each viewer participated into the event for maximum an hour.

\begin{figure}[t] \centering
\begin{tabular}{cc}
\includegraphics[width=0.48\columnwidth,height=3.5cm]{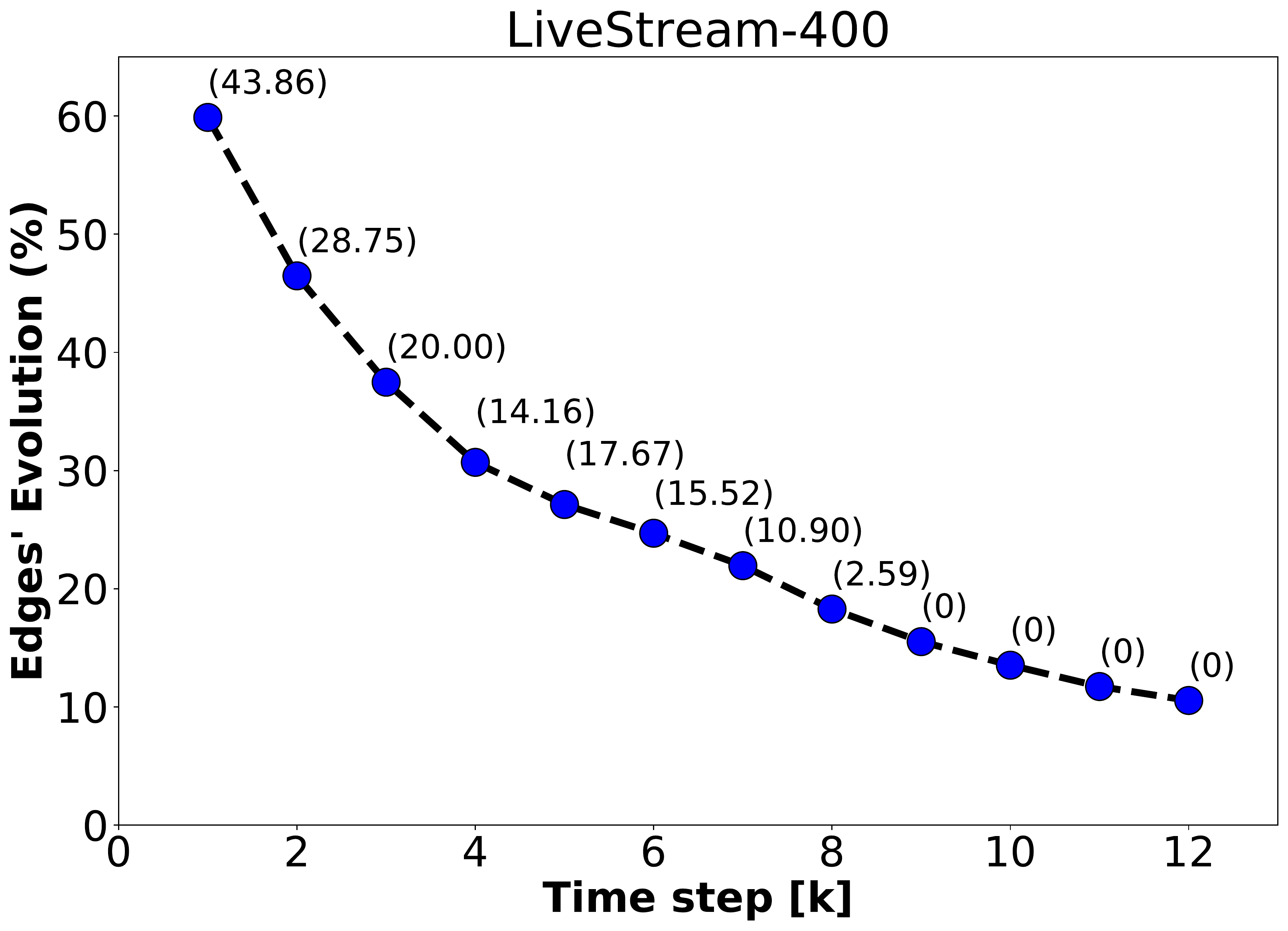} &
\includegraphics[width=0.48\columnwidth,height=3.5cm]{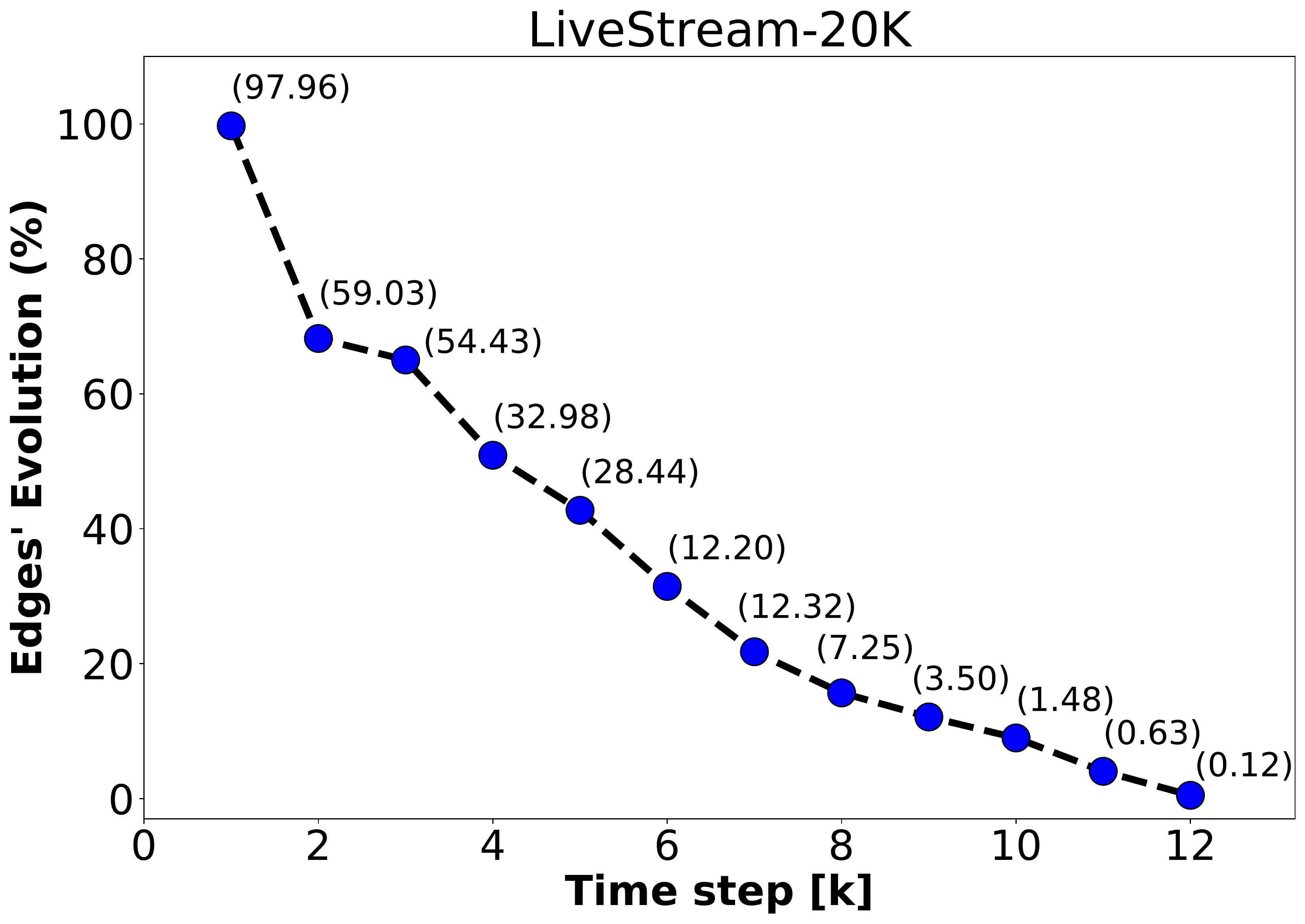}
\\
(a) & (b)
\end{tabular}
\caption{Edge's evolution over the graph snapshots in (a) LiveStream-400 and (b) LiveStream-20K. In the parentheses we denote the respective nodes' evolution.} \label{fig:EdgeEvolution}
\end{figure}

In Figure~\ref{fig:EdgeEvolution} we report the edges' evolution between two consecutive time steps. The edges' evolution is defined as $(1 - \frac{|\mathcal{E}_{k-1} \cap \mathcal{E}_{k}|}{|\mathcal{E}_k|})\cdot 100$. If the edges' evolution is equal to 100\%, then all edges have been changed when the graph has evolved from time step $k-1$ to $k$. Similarly, we compute the nodes' evolution, reported in the parentheses of Figure~\ref{fig:EdgeEvolution}. We can observe that in both datasets, the edges and nodes significantly change at the first time steps. This occurs because the majority of the viewers emerged at the beginning of the video streaming events. At the last time steps, where nodes are preserved over consecutive snapshots, the edges' evolution in both datasets is in the range of 5-10\%.

\subsection{Evaluation Protocol}

We evaluate the performance of our model in the link prediction task on the generated graphs. In our experiments we train our model on the graph snapshots until the time step $k$, to predict the unobserved edges that will occur in the next graph snapshots $\mathcal{G}_{k+1}$, $\mathcal{G}_{k+2}$, $\ldots$, $\mathcal{G}_K$, that is all the connections/links that the viewers/nodes will establish in the next time steps. For each examined model at time step $k$, we generate a test set with the unobserved edges in the next time steps, denoted by $\mathcal{O}_k = \mathcal{E}_K \backslash \mathcal{E}_k$, where $\mathcal{E}_K$ is the set of all the edges in all the $K$ graph snapshots $\mathcal{G}_1$, $\mathcal{G}_2$, $\ldots$, $\mathcal{G}_K$. Following the evaluation protocol of \cite{hamilton17,Pareja2020}, we learn the node representation matrix $\mathbf{Z}_k$ for each time step $k$, and calculate the weight of an unobserved connection/link $o(u,v) \in \mathcal{O}_k$ by concatenating the representations $\mathbf{Z}_k(u)$ and $\mathbf{Z}_k(v)$, which are then fed into a Multi-Layer Perceptron (MLP). We evaluate our experiments in terms of Mean Absolute Error ($MAE$) and Root Mean Square Error ($RMSE$), which are defined as follows: 

$$MAE= \frac{\sum_{o(u,v) \in \mathcal{O}_k} \left| \mathbf{Z}^\top_k(u) \mathbf{Z}_k(v) - A_{k} (u,v)   \right|}{|\mathcal{O}_k|}$$

$$RMSE = \sqrt{\frac{1}{|\mathcal{O}_k|}\sum_{o(u,v) \in \mathcal{O}_k} {\bigg(\mathbf{Z}^\top_k(u) \mathbf{Z}_k(v)  - A_k(u,v)\bigg)^{2}}}$$ 
Note that RMSE emphasizes more on larger errors on the link prediction task than the MAE metric.

\subsection{Compared Methods}
In our experiments we examine the performance of the following models:
\begin{itemize}
    \item \textbf{GraphSage}\footnote{\url{https://github.com/williamleif/GraphSAGE}} \cite{hamilton2017} is a static graph representation learning strategy that aggregates existing node representations to generate representations for unobserved nodes via Random Walks.
    \item \textbf{DynVGAE}~\cite{mahdavi2019} is a dynamic Joint-Variational AutoEncoder approach that shares weights between consecutive variational graph AutoEncoders. As there is no available implementation of DynVGAE, we implemented it from scratch and we made the source code publicly available\footnote{\url{https://github.com/stefanosantaris/DynVGAE}}.
    \item \textbf{EvolveGCN}\footnote{\url{https://github.com/IBM/EvolveGCN}}~\cite{Pareja2020} is a dynamic approach that employs recurrent models between GCNs to capture the graph evolution.
    \item \textbf{DySAT}\footnote{\url{https://github.com/aravindsankar28/DySAT}}~\cite{Sankar2020} is a deep neural network approach with stacked self-attention layers to capture the graph evolution.
    \item \textbf{VstreamDRLS} is the proposed model. For reproduction purposes we made our source code publicly available.
\end{itemize}
For all the examined models we tuned the hyper-parameters based on a grid selection strategy and report the results with the best configuration. In Section \ref{sec:param} we study the influence of the window size $w$ and the dimension $d$ of the node representations on the performance of each examined model.

\subsection{Performance Evaluation} \label{sec:evaluation}

\begin{figure}[t] \centering
\begin{tabular}{cc}
\includegraphics[width=0.46\columnwidth,height=3.5cm]{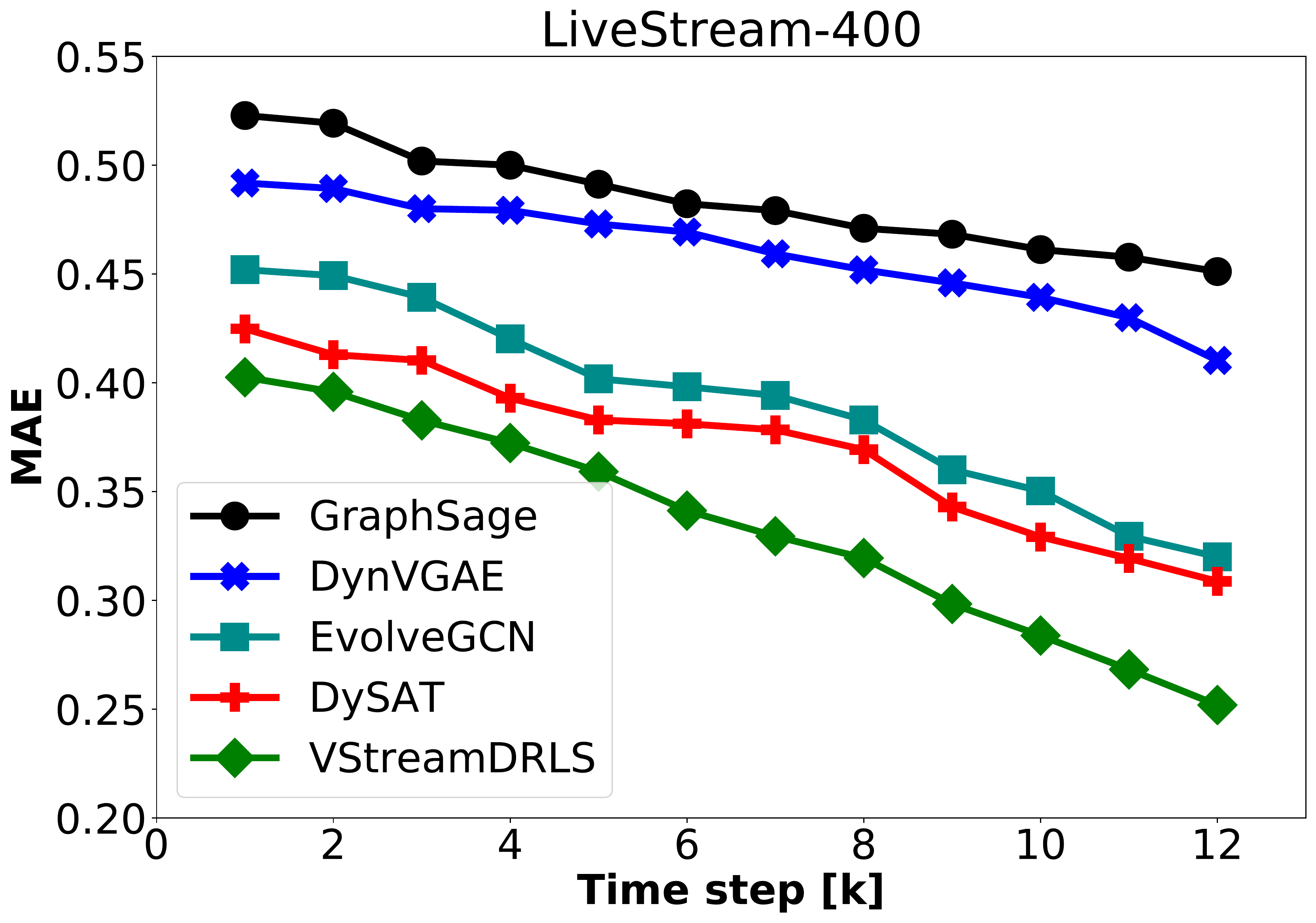} &
\includegraphics[width=0.46\columnwidth,height=3.5cm]{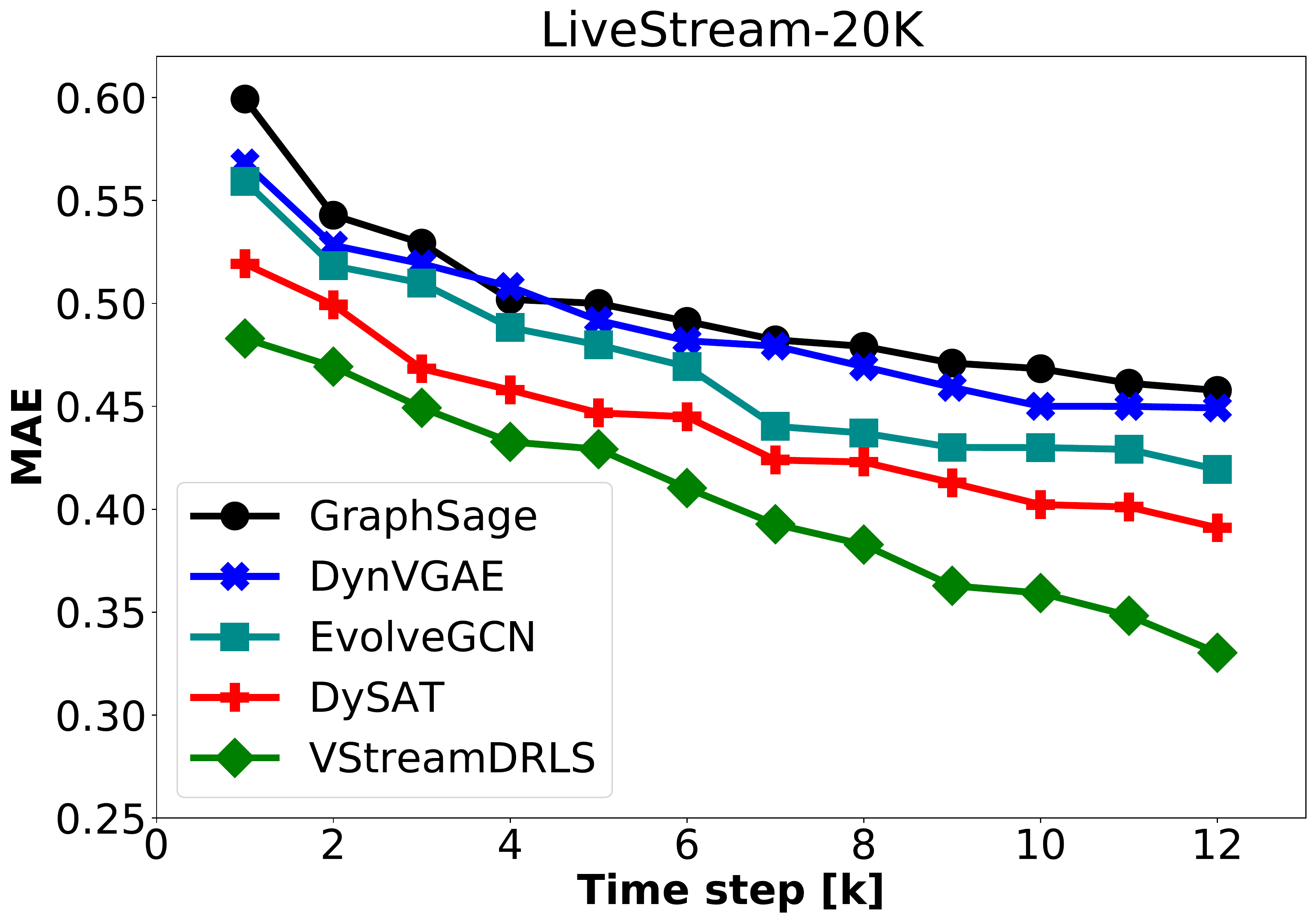}
\\
(a) & (b) \\ \\

\includegraphics[width=0.46\columnwidth,height=3.5cm]{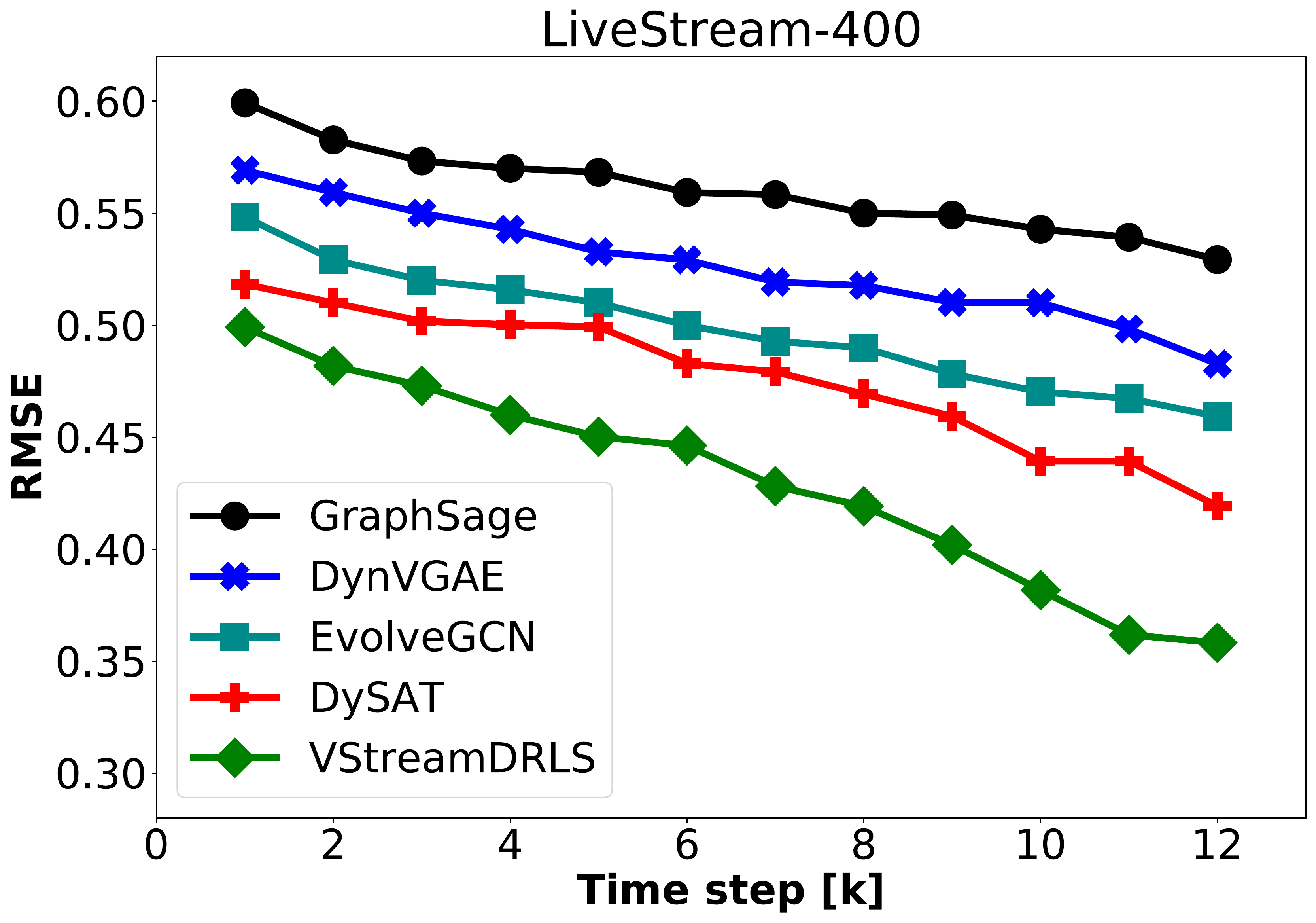} &
\includegraphics[width=0.46\columnwidth,height=3.5cm]{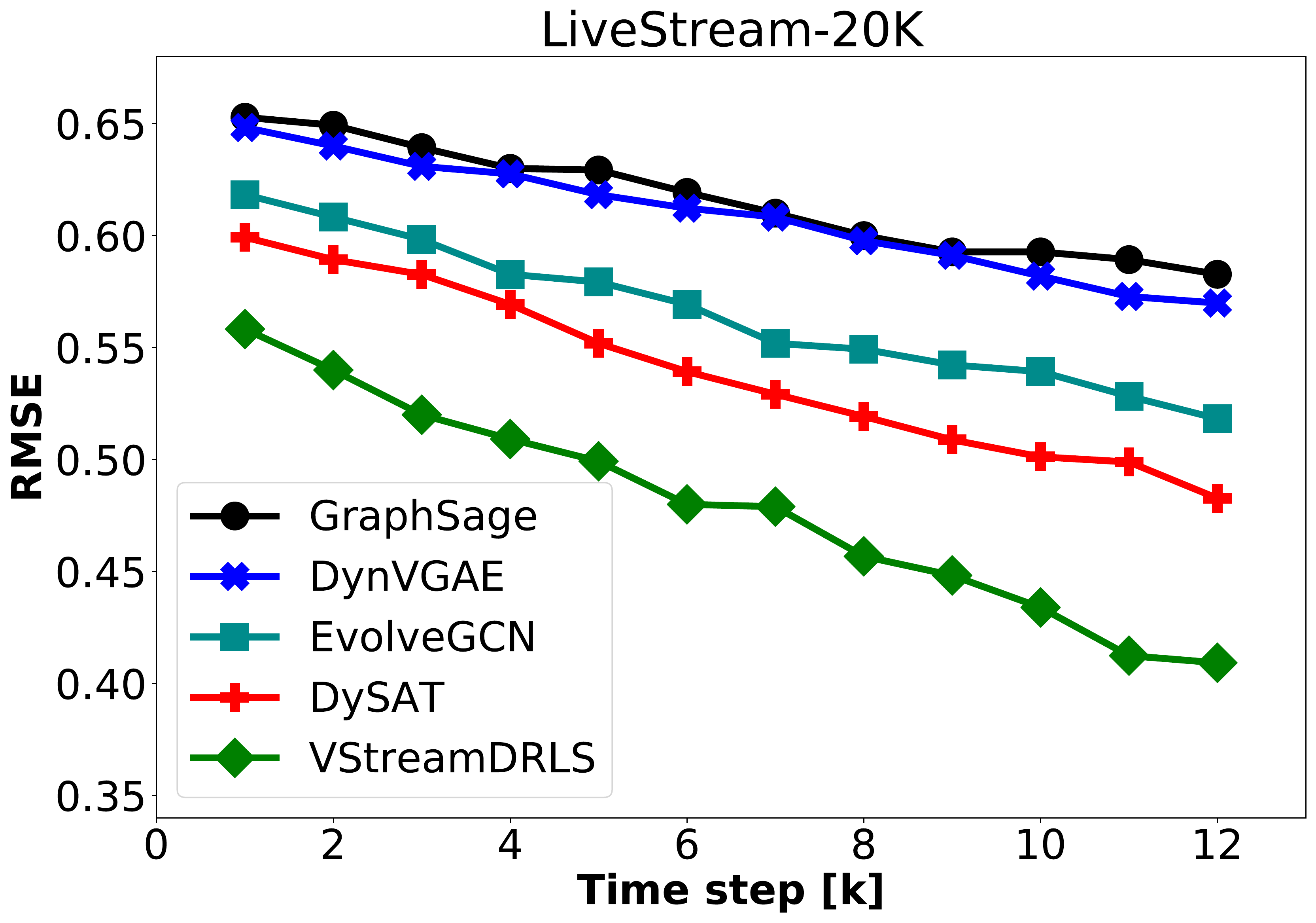}
\\
(c) & (d)

\end{tabular}
\caption{Performance evaluation in terms of (a)-(b) MAE and (c)-(d) RMSE in the LiveStream-400 and LiveStream-20K datasets} \label{fig:link_eval}
\end{figure}

In Figure \ref{fig:link_eval} we evaluate the performance of the examined models in terms of MAE and RMSE. We observe that all models achieve a lower prediction error in the LiveStream-400 dataset than in the LiveStream-20K dataset. This occurs because the LiveStream-400 graph is more dense than the graph generated in LiveStream-20K. Due to the less sparse connections/links over the time steps in LiveStream-400, the examined graph representation learning approaches achieve higher performance than in the LiveStream-20K dataset. Compared to the dynamic approaches, GraphSage underperforms in both datasets, as GraphSage learns node representations from static graphs ignoring the graph evolution. This indicates that capturing the graph evolution over the time steps has a significant impact on the performance of the examined models in the link prediction task.

The VStreamDRLS model outperforms the baseline strategies demonstrating its ability to better capture the graph evolution when predicting the unobserved links. The second best model is DySAT in both evaluation datasets. DySAT utilizes a self-attention mechanism to generate node representations via stacked self-attention layers, aiming to capture periodical patterns directly on the learned node representations. Instead, in our model we perform self-attention on the GCN parameters, that is on the weights between consecutive GCNs. In doing so, VStreamDRLS attends the graph evolution and makes the GCN of the last time step learn more accurate node representations via the convolutional layers in a live video stream event, where the graph snapshots significantly differ between consecutive time steps. Therefore, compared to DySAT our proposed model achieves $11.8$ and $9.1\%$ relative drops in terms of MAE and RMSE, respectively in the LiveStream-400 dataset. Similarly, the relative drops in  LiveStream-20K are $11.25$ and $11\%$ in terms of MAE and RMSE, respectively. As RMSE emphasizes more on larger prediction errors than MAE, the relative drop of the RMSE metric is higher in the LiveStream-20K dataset than in LiveStream-400. This occurs as the graph in LiveStream-20K evolves more frequently than the graph in  LiveStream-400, as illustrated in Figure~\ref{fig:EdgeEvolution}. Note that live video streaming events produce graphs that frequently evolve over the time steps. Our proposed model still outperforms the baselines in the LiveStream-20K dataset, reflecting the relatively high performance of our model on the real-world setting.

\subsection{Parameter Analysis}\label{sec:param}

Next, we perform a parameter sensitivity analysis of the examined models. We first study the effect of the window size $w$, that is the number of the previous graph snapshots that are used to train the examined models. Therefore, at each time step $k$, with $k=1,\dots,K$, the examined models exploit the available information at the graph $\mathcal{G}_k$ and the $w$ previous graph snaphots. We vary the window size $w$ from $1$ to $5$ by a step of $1$. In this set of experiments we report the average RMSE over all the time steps for each model. Note that GraphSage is a static graph representation learning approach which ignores the previous graph snapshots. Therefore, we omit GraphSage from this parameter analysis.

\begin{figure}[t] \centering
\begin{tabular}{cc}
\includegraphics[width=0.46\columnwidth,height=3.5cm]{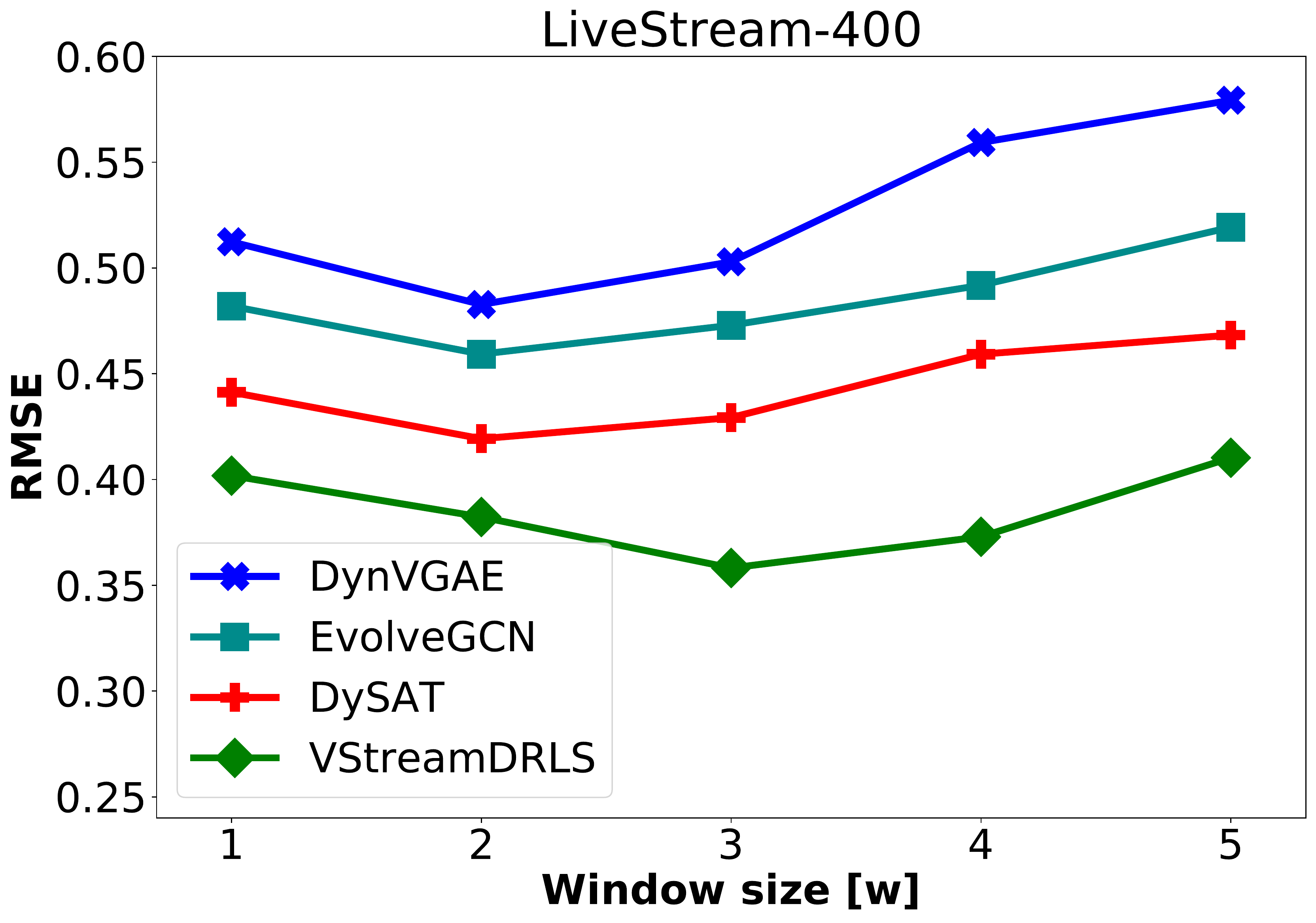} &
\includegraphics[width=0.46\columnwidth,height=3.5cm]{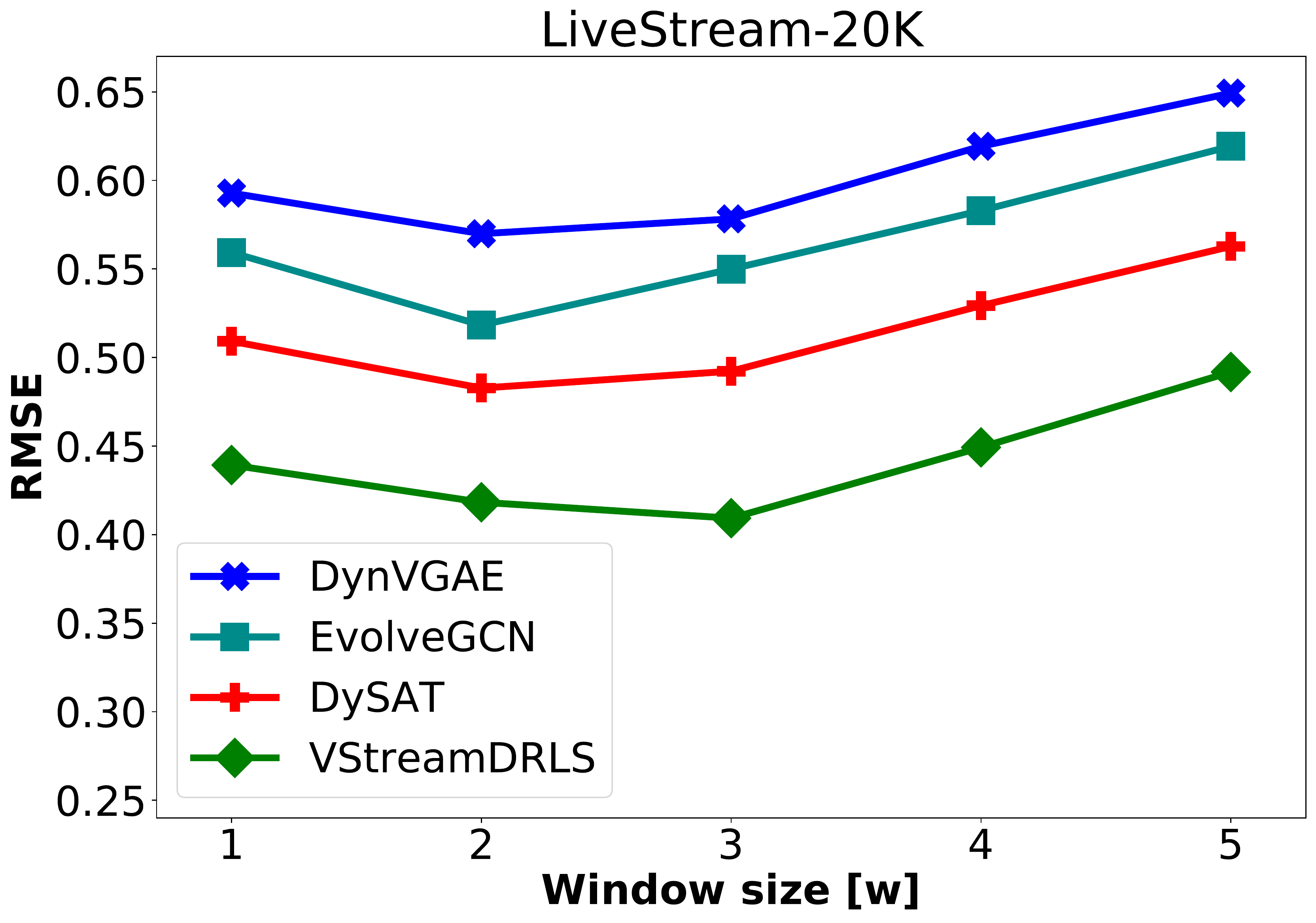}

\\
(a) & (b) \\ \\

\includegraphics[width=0.46\columnwidth,height=3.5cm]{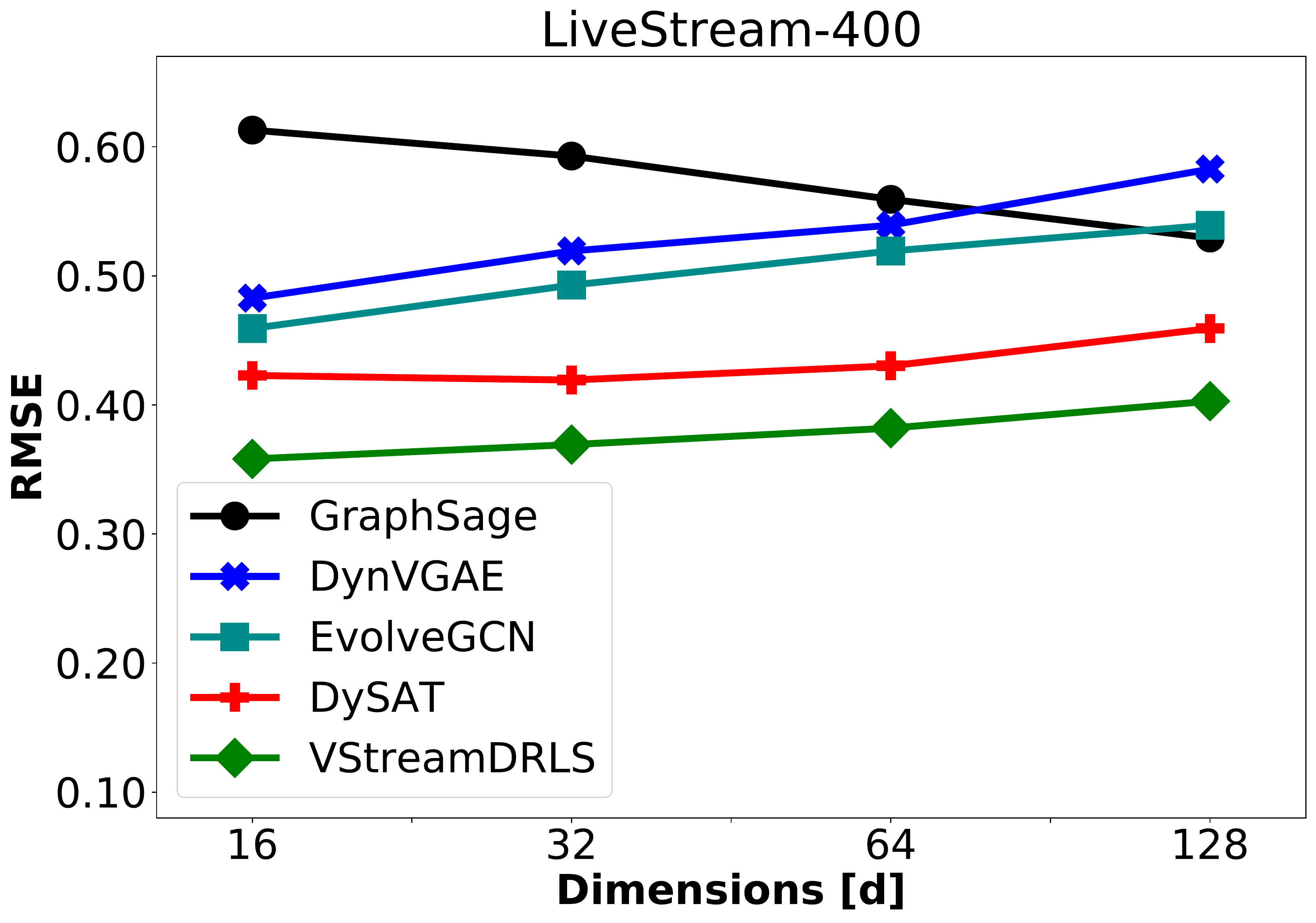} &
\includegraphics[width=0.46\columnwidth,height=3.5cm]{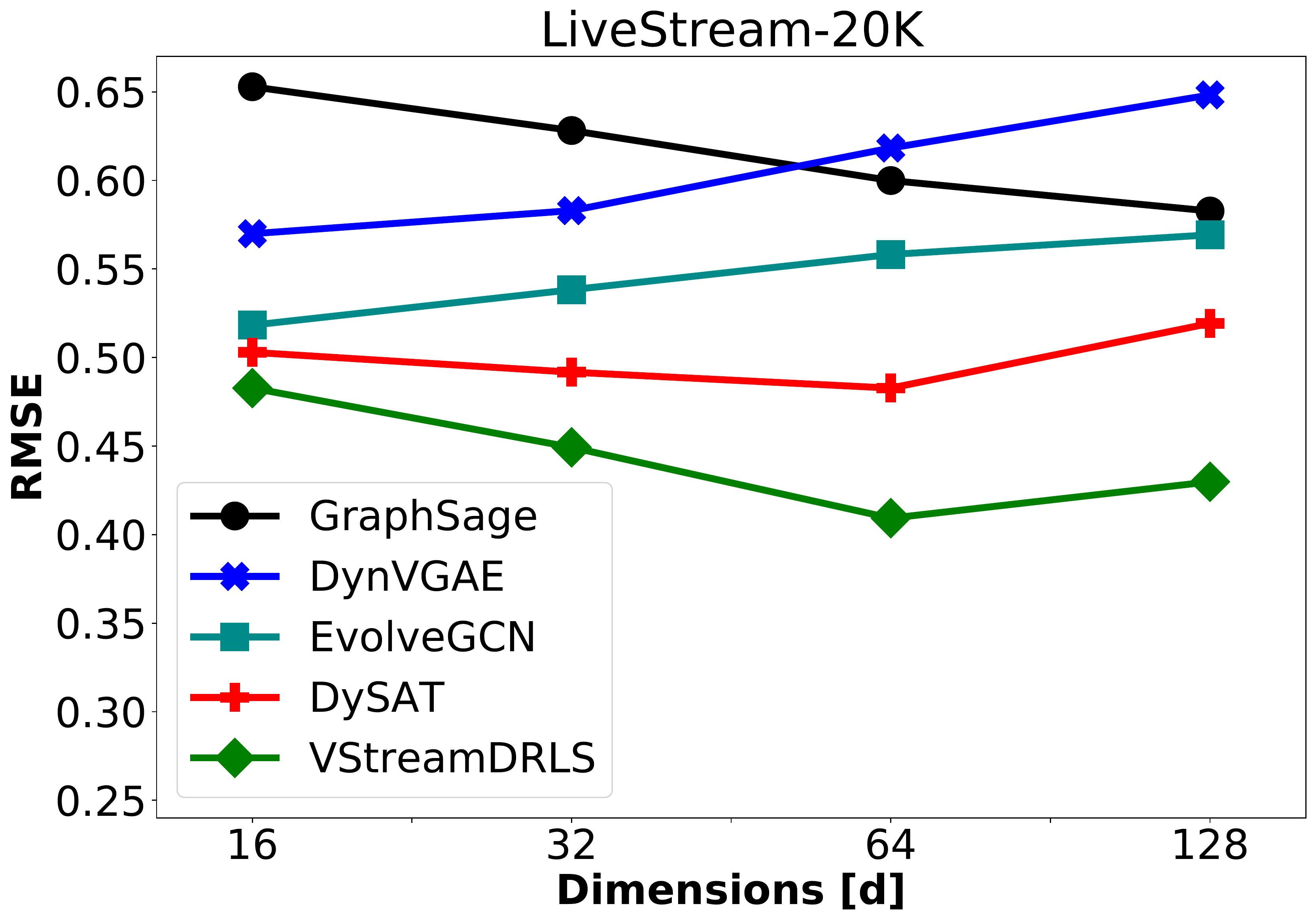}
\\
(c) & (d)

\end{tabular}
\caption{Effect on RMSE when varying (a)-(b) the time window size $w$ and (c)-(d) the number of dimensions $d$ of the generated node representations.} \label{fig:ParameterAnalysis}
\end{figure}

On inspection of Figures \ref{fig:ParameterAnalysis} (a)-(b) we observe that all the baseline approaches achieve lower RMSE when they learn node representations using $w=2$ previous graph snapshots. Increasing the window size to $w>2$ negatively affects the performance of the baselines, as more graph snapshots introduce noise to the learning process of their models, provided that in both evaluation datasets the generated graphs significantly change over consecutive time steps. In addition, decreasing the window size to $w=1$ prevents the baseline models from capturing the evolution of the graph more accurately. Instead, the proposed VStreamDRLS model achieves the best performance when the window size is equal to $w=3$ for both datasets. Setting a larger time window $w$ in our model than in the baselines indicates that our VStreamDRLS model is able to filter out the noise of the fast evolving nodes/edges over the graph snapshots. As a consequence the proposed model generates more accurate node representations than the baselines. As explained in Section \ref{sec:VstreamDRLS}, this happens because our self-attention mechanism captures the evolution of the graph via the weights of the GCN parameters. 

In the next set of experiments presented in Figures \ref{fig:ParameterAnalysis} (c)-(d), we examine the influence of the number of dimensions $d$ of the generated node representations $\mathbf{Z}_k$. We observe that GraphSage requires a high number of dimensions to achieve a relatively high performance, however, significantly increasing the computational cost when learning the model. In the LiveStream-400 dataset, the proposed VStreamDRLS model requires a lower number of dimensions than the second best method DySAT, that is $d=16$ dimensions for VStreamDRLS and $d=32$ dimensions for DySAT. In the LiveStream-20K dataset, the best configuration for both VStreamDRLS and DySAT is when setting $d=64$. Thus, both the size of the graph and the vastly evolving behavior in the LiveStream-20K dataset require a higher number of dimensions of the node representations than in LiveStream-400.

\subsection{Discussion}

Summarizing, our model consistently outperforms the baseline approaches in both evaluation datasets. In addition, the proposed VStreamDRLS model learns node representations using a higher window size $w$ than the baselines. The main challenge of considering more graph snapshots (larger values of $w$) is to filter out the noise of the fast evolving nodes/edges in a live video streaming event. This is achieved by our model based on the proposed self-attention mechanism to transfer the weight parameters between consecutive GCN models. As a consequence, our model removes the noise during the learning process of the node representations when the graphs significantly evolve. 

Distributing high-quality live video content is a network demanding process with enterprise networks having several bandwidth limitations. Therefore, accurately predicting the network capacity between viewers is essential for  distributing the video streaming content. This means that achieving low link prediction errors in terms of MAE and RMSE may offer to enterprises a solution to improve the user experience by providing content with significantly high resolution, such as $4K$ videos \cite{Taghouti2016}. Instead, graph representation learning models that underperform and provide less accurate network capacity predictions have negative impact on the performance of video streaming, as the video content will be erroneously distributed to viewers at different offices and increase the network traffic  \cite{Liu2020,Mahini2018,Terelius2018,Zhang2019}. Provided that the duration of a real-world live video streaming event lasts an hour on average, an accurate dynamic graph representation learning model plays an essential role in the distribution of a live video streaming event. In practice, an accurate model can significantly reduce the required time for each viewer to discover and establish connections with other viewers, so as to efficiently distribute the live video streaming content. Moreover, the proposed VStreamDRLS model is a video streaming technology that can distribute the video content without a prior knowledge of the enterprise network. Finally, our solution complies with the GDPR, as it predicts the network capacity without using any personal information such as the viewers' IP addresses.

\section{Conclusions}\label{sec:conc}

In this study, we presented a dynamic graph representation learning model for live video streaming technologies on large enterprise networks, namely VStreamDRLS. To the best of our knowledge we are the first who formulated the distribution of live video streaming events as a problem of dynamic graph representation learning. A key factor of our model is to design a self-attention mechanism and transfer the weight parameters between consecutive GCNs, so as to capture the graph evolution and produce accurate node representations. Our experimental evaluation demonstrates the superiority of our model over other state-of-the-art strategies in two real-world datasets, generated by real live video streaming events on enterprise networks. Moreover, we showed that VStreamDRLS can leverage information from more historical graph snapshots than the baselines strategies, indicating that our model is capable of efficiently filtering out the noise, produced by consecutive graph snapshots with significant differences. Finally, for reproduction purposes both datasets and the implementation of our model are publicly available. An interesting future direction is to study the performance of the proposed model on evolving social networks, by taking into account how users emerge and establish connections over time.

\bibliographystyle{IEEEtran}
\bibliography{IEEEexample}

\end{document}